\documentclass[runningheads]{llncs}
\usepackage{graphicx}
\usepackage{listings}
\usepackage{todonotes}
\usepackage{hyperref}
\usepackage{xcolor}

\usepackage[framemethod=tikz]{mdframed}
\definecolor{myblue}{rgb}{0.122, 0.435, 0.698}
\definecolor{mygreen}{rgb}{0.125, 0.525, 0.220}
\definecolor{myyellow}{rgb}{0.588, 0.439, 0.000}
\newmdenv[innerlinewidth=0.5pt,roundcorner=4pt,innerleftmargin=6pt,
          innerrightmargin=6pt,innertopmargin=6pt,innerbottommargin=6pt,
          linecolor=myblue,backgroundcolor=myblue!25!white]{mybluebox}
\newmdenv[innerlinewidth=0.5pt,roundcorner=4pt,innerleftmargin=6pt,
          innerrightmargin=6pt,innertopmargin=6pt,innerbottommargin=6pt,
          linecolor=mygreen,backgroundcolor=mygreen!25!white]{mygreenbox}
\newmdenv[innerlinewidth=0.5pt,roundcorner=4pt,innerleftmargin=6pt,
          innerrightmargin=6pt,innertopmargin=6pt,innerbottommargin=6pt,
          linecolor=myyellow,backgroundcolor=myyellow!25!white]{myyellowbox}

\definecolor{codegreen}{rgb}{0,0.6,0}
\definecolor{codegray}{rgb}{0.5,0.5,0.5}
\definecolor{codepurple}{rgb}{0.58,0,0.82}
\definecolor{backcolour}{rgb}{0.95,0.95,0.92}
\lstdefinestyle{mystyle}{
    numberbychapter=false,
    captionpos=b,                    
    backgroundcolor=\color{backcolour},   
    commentstyle=\color{codegreen},
    keywordstyle=\color{magenta},
    numberstyle=\tiny\color{codegray},
    stringstyle=\color{codepurple},
    basicstyle=\ttfamily,%
    breakatwhitespace=false,         
    breaklines=true,                 
    keepspaces=true,                 
    numbers=left,                    
    numbersep=5pt,                  
    showspaces=false,                
    showstringspaces=false,
    showtabs=false,                  
    tabsize=2
}
\begin{document}
\newcommand{\fullpaper}[1]{\noindent{\color{red}[#1]}}
\newcommand{\ack}{The development of the sphinx-prolog and sphinx-problog Jupyter Book plugins was supported by the TAILOR Network -- an ICT-48 European AI Research Excellence Centre funded by EU Horizon 2020 research and innovation programme, grant agreement number 952215.}
\title{%
Simply Logical -- 
The First Three Decades%
}
\titlerunning{Simply Logical -- The First Three Decades}
\author{%
Peter~Flach\inst{1}\orcidID{0000-0001-6857-5810} \and%
Kacper~Sokol\inst{1}\orcidID{0000-0002-9869-5896} \and%
Jan~Wielemaker\inst{2}\orcidID{0000-0001-5574-5673}%
}
\institute{%
Intelligent Systems Laboratory, University of Bristol, United Kingdom\\
\email{\{peter.flach, k.sokol\}@bristol.ac.uk}%
\and%
SWI-Prolog Solutions b.v., Netherlands\\
\email{jan@swi-prolog.org}%
}
\maketitle              %
\begin{abstract}
This paper charts the evolution of the Prolog textbook \textit{Simply Logical -- Intelligent Reasoning by Example} 
from print with runnable programmes on a 3.5-inch diskette published in 1994, %
via various intermediate online versions, 
to the fully interactive online edition available today. %
Our aim is to investigate -- from both the writer's and the reader's perspectives -- the potential impact of technology on educational materials. %
The three authors of this paper present three distinct and complementary points of view, which come together to shape an interactive online book that offers an immersive learning experience. %
Peter describes the philosophy behind the original book and experiences teaching from it. %
Jan demonstrates how contemporary web programming has enabled a fully interactive realisation of the book's philosophy. %
Kacper reports how different technologies can be adopted to build a versatile authoring toolkit that underpins the interactive online edition of the book. %
Collating these distinct yet coherent perspectives allows us to reflect on future opportunities for interactive Prolog teaching materials arising from the continuous development of web technologies. %
The insights we offer should be of interest to teachers of Prolog as well as those wanting to find out about the latest educational technologies.%
\end{abstract}

\section{The Paperback \hfill {\small\em(Peter Flach)}\label{sec:peter}}%

As a beginning lecturer at Tilburg University in the early 90s of the previous century I started teaching a module on Artificial Intelligence (AI), then an emerging specialist topic within the still young discipline of computer science. 
Initially using a general AI textbook~\cite{luger1993artificial} I started devising practical exercises and coursework assignments in Prolog, which I had come across a couple of years before. These grew into a set of lecture notes that eventually became the \textit{Simply Logical -- Intelligent Reasoning by Example} book~\cite{flach1994simply}. I ended up publishing with John Wiley after I had a chat with a representative at a conference; they were very helpful and even suggested the title. I was fortunate in finding Bob Kowalski willing to write a foreword, in which he very graciously wrote that my book relieved him of the temptation to write a revised edition of his own book, \textit{Logic for Problem Solving}~\cite{kowalski1974logic}. 

Rather than trying to cover all of AI, which even then had grown into a diverse field, I decided to focus on reasoning in various forms: deductive, inductive, reasoning with natural language, reasoning with incomplete information, etc., which approach is captured by the ``intelligent reasoning'' part of the book's title.
The ``by example'' moniker reflected the didactic philosophy: teaching by showing, learning by doing.
Every AI technique discussed was accompanied by a Prolog programme implementing it. These programmes served two didactic purposes. By running them, the student could get a feeling what a particular technique is all about. But perhaps more importantly, the declarative reading of each programme was an integral part of the explanation of the technique it implemented. For the more elaborate programmes, special attention was paid to their stepwise development, explaining key issues along the way. 
My hope was that the reader would mount the accompanying 3.5'' diskette into their computer, and try out the Prolog programmes alongside reading the book. While this was perhaps optimistic back then, the online format available today integrates the ``showing'' and ``doing'' sides of the educational coin very well, as will be demonstrated below.

I still use \textit{Simply Logical} in my lectures as an introduction to both logic programming and AI. Treating the subjects side by side allows for some pretty powerful examples, such as a resolution prover for full clausal logic as a demonstrator for \href{https://book.simply-logical.space/src/text/2_part_ii/5.3.html}{breadth-first search}. %
Rather than modernising the text I have opted for modernising the demonstrators and programming assignments. 
For example, the natural-language question-answering programme described in \href{https://book.simply-logical.space/src/text/3_part_iii/7.3.html}{Chapter~7} of the book, %
which nowadays we would call a chatbot, was developed into an \href{https://github.com/So-Cool/prolexa}{Amazon Alexa skill}. %
This demonstrates the power of a hybrid system consisting of speech-to-text and text-to-speech components enabled by deep learning, and a logical engine in the middle capable of explainable logical reasoning. 
Other material developed over the years includes a \href{https://github.com/COMS30106/assignment/wiki/Assignment-2}{Prolog-driven agent} that needs to solve a logical puzzle while moving around in a grid world %
and a \href{https://revealjs.com/}{Reveal.js} \href{https://labs.simply-logical.space/}{slide deck} that can be used for self-study and trying out Prolog programmes without leaving the browser. %
The latter is based on SWISH, which Jan introduces in the following section.

\section{Interactive Prolog in the Browser \hfill {\small\em(Jan Wielemaker)}\label{sec:jan}}

Crucial for realising the fundamental philosophy behind \textit{Simply Logical} is the ability to
run code interactively in the browser. Modern browser and service
technology allows for several designs:
\begin{enumerate}
\item Have Prolog in a sandbox (such as a virtual machine or container) on a server
  and provide ways to submit a programme and view its output. An
  example of this approach can be found at
  \href{https://www.tutorialspoint.com/execute\_prolog\_online.php}{Coding
    Ground}.
\item Have a web service with dedicated support for Prolog that
  offers a sandboxed Prolog execution environment. SWISH\footnote{\url{https://swish.swi-prolog.org/}} -- \textit{SWI-Prolog for SHaring}~\cite{wielemaker2015swish} -- is implemented based on this paradigm. %
\item Execute Prolog in the browser. Some Prolog systems are written
  directly in JavaScript, thus enabling them to be run in the browser, e.g.,
  \href{http://tau-prolog.org/}{Tau-Prolog}.
  Another solution is to compile Prolog to JavaScript~\cite{morales2012lightweight}.
  Alternatively, Web Assembly (WASM) offers a low-level
    virtual machine for the browser, which
    allows running
    more traditional C-based Prolog implementations; %
    for example, \href{https://emscripten.org/}{Emscripten} provides a Clang-based
    compiler suite that targets WASM, underpinning implementations such as %
    the \href{http://ciao-lang.org/playground}{Ciao
      Playground} and
    \href{http://dev.swi-prolog.org/wasm/shell}{SWI-Prolog WASM port}.
\end{enumerate}
The first category typically provides a batch-style
interaction, making it difficult to adapt for (online and interactive) educational resources; %
the latter two approaches are more suitable in this respect.

SWISH,
originally developed by Torbj\"orn Lager, is based on
\href{https://www.swi-prolog.org/pldoc/package/pengines}{Pengines} -- an
architecture to access Prolog on a remote server --
and consists of
a monolithic multi-threaded SWI-Prolog server. Its user interface is written in
JavaScript with server-side assistance for semantic highlighting,
feedback on predicates, template insertion and other user-focused features.
Pengines underlies Prolog execution, which
takes place in a sandboxed environment that consists of
a Prolog thread and a temporary module. Executing a query (1) transfers
the code and query to the server; (2) starts a thread and creates a
temporary module; (3) loads the code into the module; and (4) executes
the code. This operationalisation allows for rich interaction with the client, e.g.,
``read'' from the client, ask for alternative solutions or trace
through the programme in a manner similar to traditional Prolog tracers. Other
noteworthy features of SWISH include:
\begin{itemize}
\item The ability to store programmes on the server under Git-style version %
  management. Programmes can either be private or visible to the public, %
  and may \textit{include} each other.
\item Availability of \textit{render plugins} that allow rendering Prolog terms -- e.g., as tables or graphics that represent chess boards, charts and graphs
-- using modern web browser technology.
\item An \href{https://www.r-project.org/}{R Project} interface
  offering access to R's algorithms and chart outputs.%
\item \textit{Notebook} format allowing for rich interactive
  pages that provide a sequence of HTML, Markdown,
  programme and query cells.%
\end{itemize}

SWISH has proved to be an invaluable resource for the Prolog community, including educators.
It facilitates (Prolog) learning, code discussion and exchange as well as %
shared access to data for data science.
After its launch in 2014 the
system quickly gained popularity as an alternative to local
installations of Prolog systems used for exercises in Prolog classes at
schools worldwide.
As of December 2022, it hosts over 150,000 programmes and nearly %
15,000 notebooks, many of which in several versions.
It sustains on
average hundreds of concurrent users, peaking at over one thousand %
users, usually towards the end of each academic cycle. %
Notably, teachers started adapting their
educational materials and course workflows to SWISH for more than just
a convenient online Prolog execution environment. %
Some wrote course material as SWISH notebooks.
Others use SWISH notebooks for exercises;
they organise the content as questions, background data,
example queries and, sometimes, skeleton or partial solutions. %
A student then \textit{forks} the notebook, completes it,
saves it and sends the link with the completed notebook to the teacher.

As interactive in-browser programming exercises were gaining in
popularity, I began to wonder whether we could do something similar
to teach Prolog as a programming language.
With \href{http://www.let.rug.nl/bos/lpn//index.php}{\emph{Learn Prolog Now!}}~\cite{blackburn2001learn}
being one of the more popular online (and later paperback~\cite{blackburn2006learn}) Prolog books, I
experimented with making it interactive.
\textit{Learn Prolog Now!} is written in \LaTeX{}, which is then converted to
an \emph{unintelligible} HTML code used to publish the online version. %
Because of that, I built the %
\href{https://lpn.swi-prolog.org/}{interactive version}~\cite{blackburn2014learn} by using a
\textit{rewriting proxy server}.
This is a Prolog programme that:
\begin{enumerate}
  \item uses SWI-Prolog's HTML parser and generator to parse the original pages;
  \item identifies the code blocks that represent programmes and queries;
  \item re-writes the programme fragments into new HTML elements
that connect each programme to relevant queries, allowing these pairs to
be transformed into an embedded SWISH instance; and
  \item inject JavaScript into the page to make the page -- specifically, the code boxes -- interactive.
\end{enumerate}
While side-stepping the need to manually edit the book source, %
this automation is far from perfect.
Observing this particular use case of SWISH, nevertheless, motivated Peter and Kacper to make %
\textit{Simply Logical} similarly interactive.
They decided to
manually identify the programmes and queries as it was clear
that given the available technology it is impossible to do so programmatically with
acceptable precision.%

\section{The Online Edition(s) \hfill {\small\em(Kacper Sokol)}\label{sec:kacper}}%

The book was first made freely available in 2007, when the copyright reverted back to Peter, who decided to release its PDF and associated programmes online through a \href{http://people.cs.bris.ac.uk/~flach/SimplyLogical.html}{dedicated web page}. %
In 2015, as a pilot, most of the original book source -- written in a now obsolete version of MS Word -- was ported to a collection of messy HTML files replete with superfluous markup;
the figures were extracted as non-standard EMZ files and later converted to the more modern Scalable Vector Graphics (SVG) format. %
This provided a canvas for transforming a static web page into an online, interactive resource using SWI-Prolog's SWISH platform.

\begin{figure}[!t]
    \includegraphics[width=\textwidth]{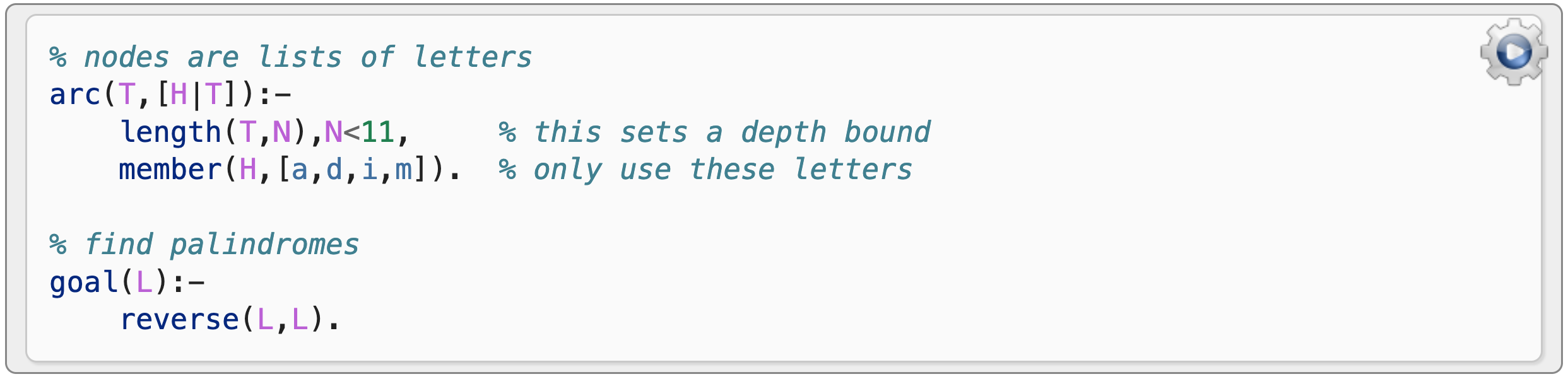}%
    \\ \ \\
    \includegraphics[width=\textwidth]{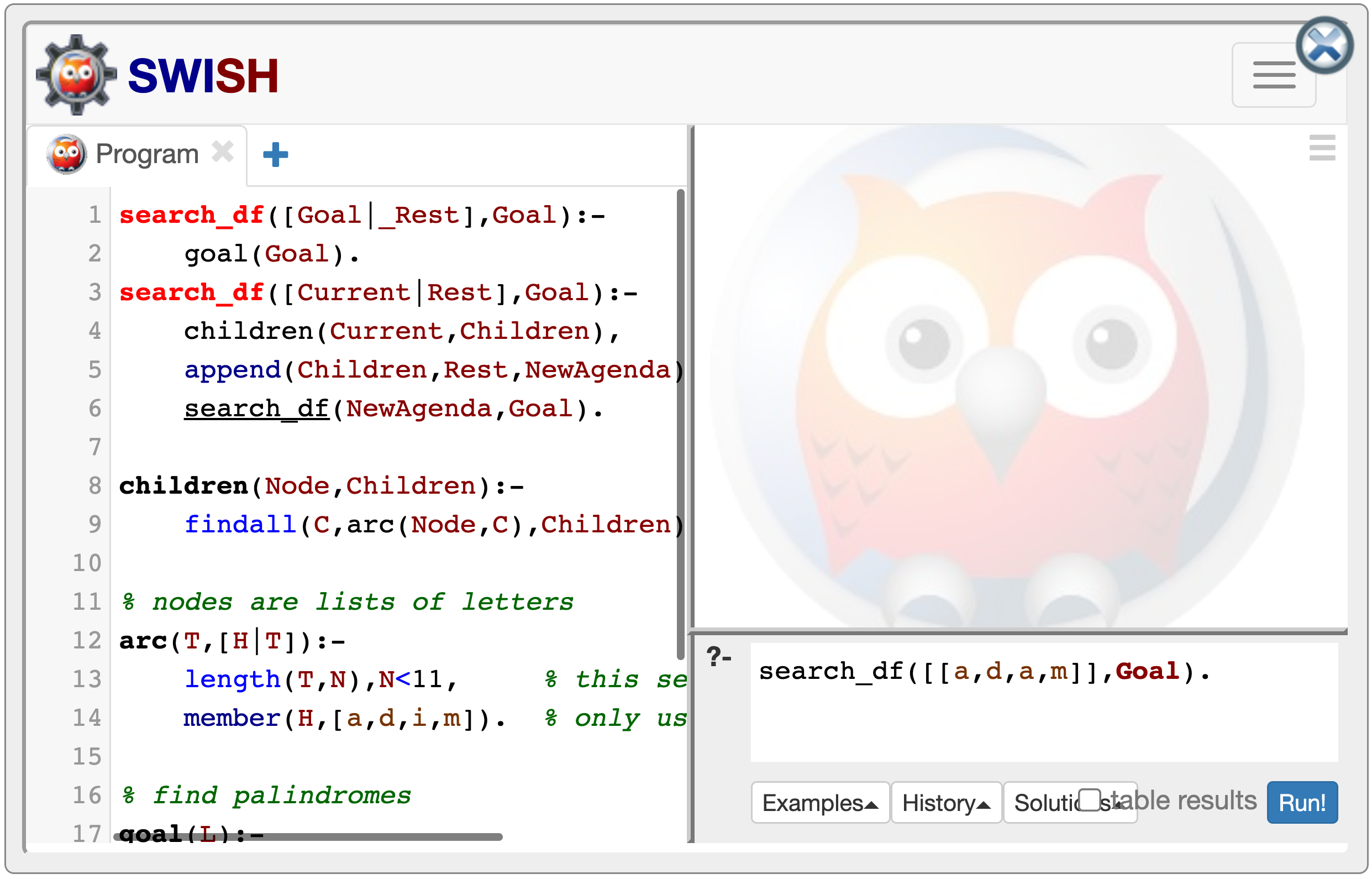}%
    \caption{Interactive SWI-Prolog code box based on SWISH. A code listing (top) can be turned into an interactive code box (bottom) by pressing the ``play'' button in its top-right corner.\label{fig:codebox}}
\end{figure}

From then on, the manual process of identifying Prolog code listings and their associated queries took place, with appropriate HTML code being embedded by hand to display interactive SWISH code boxes. %
Each SWISH box initially renders the programme as text, as in the original book.
Opening the box reveals a tripartite widow with an editable code listing (pulling in additional code as required), pre-populated Prolog queries, and an answer pane as shown in Figure~\ref{fig:codebox}. %
Two decades after publication of the original book, technology had caught up sufficiently to support the envisaged didactic philosophy, which was an almost revelatory experience. %
The \emph{first} release had thus seen the light of day.%

The script responsible for rendering SWISH code boxes was elaborate enough to streamline certain tasks.
Prolog queries displayed in the text were assigned unique IDs, which allowed them to be dynamically imported into relevant code boxes upon launch. %
Both interactive programmes and static code listings could also be tagged, providing an opportunity to import code into interactive boxes at run time. %
To give the book a fresh look and feel and prepare its presentation for a range of modern devices -- desktop and laptop computers, tablets and mobile phone screens -- the HTML content was adapted to the \href{https://getbootstrap.com/}{Bootstrap} framework. %
The book was split into four HTML files: Part~I, Part~II, Part~III and Appendix, complemented with an index page, with the source hosted on GitHub\footnote{\url{https://github.com/simply-logical/simply-logical/}}. %
The \emph{second} release of the book's online version was built with \href{https://jekyllrb.com/}{Jekyll} to generate static web pages, which were hosted on GitHub Pages under the dedicated \url{https://book.simply-logical.space/} domain.%

However, the main text was now written in HTML, which was still fairly limiting, especially when it comes to authoring (new or derived) content.
With Markdown being much simpler to write, edit and maintain as well as more human-readable and covering all of the markup needs, I started porting and cleaning the source.
In the process, the diversity of the book's building blocks -- text, figures and code listings -- sparked an idea for a bespoke content generation system where the resources are split into prime elements that can be reused to compose new material.
Initially, this novel workflow relied on Jekyll, which proved to be limiting and prompted me to embark in late 2018 on developing a bespoke content generation platform.
Its prototype, while versatile, required the source documents to be extensively annotated, making the authoring experience overly complicated and stalling the project.%

When Jupyter Book~\cite{jupyter2020book} arrived in early 2020 it rekindled my hope for a modular version of \emph{Simply Logical}. %
The platform allowed composing content from diverse building blocks and offered nearly all of the required functionality, apart from support for SWISH code boxes. %
While based on the Sphinx Python documentation generation engine, it nonetheless could be extended with a custom plugin. %
This encouraged me to finalise the separation, conversion and cleaning of the book content, and to begin developing the necessary Jupyter Book extensions. %
The result is a highly modular design, where the book is split into a collection of files based on their type: text, exercise, solution, code and figure. %
All of these materials are published on GitHub under the permissive Creative Commons (BY-NC-SA) licence, which allows easy reuse, incorporating into bespoke courses, or adapting into alternative educational resources such as practical training sessions. %
This release marked the \emph{third} instalment of the online edition~\cite{flach2022simply}.%

From a technical perspective, the latest version is realised through a collection of bespoke Jupyter Book plugins (that also work with Sphinx) spanning functionality specific to SWI-Prolog~\cite{wielemaker2012swiprolog}.
These extensions were later adapted to be compatible with the \href{https://cplint.ml.unife.it/}{cplint}~\cite{riguzzi2018foundations} and \href{https://dtai.cs.kuleuven.be/problog/}{ProbLog}~\cite{deraedt2007problog} probabilistic programming languages. %
Specifically, I developed and released sphinx-prolog\footnote{\url{https://github.com/simply-logical/sphinx-prolog/}}, which among other things allows to embed interactive SWI-Prolog and cplint code boxes based on their respective SWISH implementations. %
The sphinx-problog\footnote{\url{https://github.com/simply-logical/sphinx-problog/}} extension offers analogous functionality for ProbLog facilitated by %
the online execution environment underpinning the code examples published as part of the ProbLog website.

\begin{lstlisting}[caption=Example of SWISH code box syntax., label=code:swish, float=t,  literate={`}{\textasciigrave}1,]
```{swish} swish:0.1
---
query-text: ?-linked(a,b,X). ?-linked(a,X,b).
query-id: swishq:1.1.1 swishq:1.1.2 swishq:1.1.3
inherit-id: swish:4.5.6 swish:4.5.7 swish:4.5.8
source-text-start: 4.5.6-start
source-text-end: 4.5.6-end
hide-examples: true
---

optional :- content.

/** <examples>
?-linked(X,a,b).
*/
```
\end{lstlisting}

Including a SWISH code box is as simple as writing a specially-formatted Markdown code block, an example of which is shown in Listing~\ref{code:swish}. %
This element is assigned a unique ID (\texttt{swish:0.1}), and its content can either be provided verbatim (\texttt{optional :- content.}) or loaded from a file. %
Through its optional parameters it is possible to: %
\begin{itemize}
\item explicitly provide the query (\texttt{query-text}) and/or import it from a tagged block (\texttt{query-id}); %
\item inject a programme from another code box through its ID (\texttt{inherit-id}); %
\item prepend (\texttt{source-text-start}) or append (\texttt{source-text-end}) code from external Prolog files; and %
\item if Prolog queries are included directly in the code via a specially-formatted query block (lines 13--15 in Listing~\ref{code:swish}), hide them from the users prior to launching the code box (\texttt{hide-examples}). %
\end{itemize}
To streamline the process of creating new online interactive educational resources, I published a suite of templates for %
Prolog\footnote{\url{https://github.com/simply-logical/prolog-book-template}}, %
cplint\footnote{\url{https://github.com/simply-logical/cplint-book-template}} and %
ProbLog\footnote{\url{https://github.com/simply-logical/problog-book-template}} %
content. %
By \textit{forking} any of these repositories one can adapt static resources or build new teaching materials in a matter of minutes.%

\section{Discussion and Outlook\label{sec:outlook}}

Developing Jupyter Book plugins and building authoring environments for interactive Prolog content set us on a path to further explore alternative technologies for composing diverse (interactive) training resources. %
As a proof of principle we created a prototype of a new publishing workflow in which multiple artefacts such as online documents, slides and computational notebooks can be generated from a unified collection of source materials~\cite{sokol2021you}. %
This contrasts with much of current practice, where authoring environments are limited to individual publishing artefacts, for example, \LaTeX{} or MS Word documents, and PowerPoint or Keynote slide decks. %
While in its early stages of development, Quarto~\cite{quarto} is a notable departure from this paradigm and a substantially more powerful realisation of the envisaged workflow. %
It is a scientific and technical publishing system that
allows to author content as Markdown files or Jupyter Notebooks,
and publish it as articles, reports, presentations, websites, blogs and books in formats such as HTML, PDF, MS Word or EPUB.
Quarto supports scientific features -- e.g., equations, citations and cross-referencing -- and inclusion of dynamic content created with various programming languages.%

Regardless of the underlying technology, our ideal scenario is an authoring environment in which the author can concentrate on producing content without much (initial) regard for the delivery format.
Similar ideas have been around for a long time -- e.g., Luhmann's Zettelkasten~\cite{luhmann-zettelkasten} -- but were lacking the technology to support them and make them efficient. %
While we hope that the next 50 years of Prolog will see these ideas and opportunities come to full fruition, numerous challenges need to be overcome.
One particular bottleneck is limited interactivity of the educational resources published on the Internet, which in part is due to the need for compute resources. %
Without access to a dedicated machine -- however it is implemented in the physical world -- interacting with code examples remains constrained.%

For example, up until recently SWISH used to run on
a single server hosted by the Vrije Universiteit Amsterdam (VU).
The single process and single hardware placed in a single network made the system
highly vulnerable.
This could be addressed by
migrating the server to a
set of \textit{federated services};
while currently SWISH runs on two servers in different locations, \textit{load balancing}
is still missing.
The concept of server-side execution, nonetheless, remains problematic because it concerns a complex, non-standard service that requires significant compute resources which are in need of continuous maintenance and a sustained stream of funding. %
An alternative approach is to move towards Prolog systems running in
the browser of each individual user.

Among other things, the browser-based approach lifts the need for sandbox
containment because this security feature is already provided by the
browser. %
Currently, SWISH rejects programmes it cannot \emph{verify} to %
be safe,
which includes scripts that
perform
\textit{intractable} meta-calling.
In the future, this could be resolved by using a
process rather than a thread for implementing Pengines, and isolating
this process using an operating system sandbox.
Additionally, as it stands, SWISH is \textit{stateless}, i.e., executing a query
against a programme does not change the server state,
since
new queries are
executed in a fresh environment.
This implies that we cannot interactively
call \texttt{assert/1} in one query and use the asserted clauses in the following queries. %
Instead, we must both assert and use the clauses in the same query. %
One big advantage of avoiding permanent state, nonetheless, is making all queries %
fully reproducible.

As noted in Section~\ref{sec:jan}, there are JavaScript-based Prolog systems %
that explicitly target the web browser, in addition to more versatile WASM implementations. %
The ideal scenario would be for the Prolog community to come up with
a ``standard'' embedding interface to enable designing interactive %
web pages using multiple Prolog backends.
However, the browser control flow is inherently incompatible with traditional
single-threaded Prolog systems since
one cannot wait \textit{synchronously} for input in a browser (e.g., call \texttt{read/1}). %
Instead, one must return control to the browser and when the input
becomes available as an \textit{event}, only then Prolog may resume.
There are two ways to achieve that.
One is to run Prolog in a
\textit{WebWorker} (roughly, a browser task without a window) and make
it communicate with the browser task associated to a normal browser
window.
The other is to \textit{yield} from the Prolog virtual
machine and \textit{resume} Prolog when the data become available.
This leads to at least two interface designs:
\begin{itemize}
    \item with a WebWorker we get
interaction with a Prolog process that is not very different from
Prolog running on a remote server; whereas
    \item using a pure JavaScript or Prolog
with \textit{yield} allows running Prolog on the same page and having
bi-directional calls between Prolog and JavaScript, which facilitates
managing the browser directly from Prolog.
\end{itemize}

Running Prolog in the browser rather than remotely on a server solves
several problems.
With Prolog being served as a static file, storage
and chat services -- which are currently part of SWISH -- can be offloaded to existing %
cloud services.
This creates a reliable distributed infrastructure
that can be sustained by several Prolog implementations.
Such diversification is important as
due to the
use of many features that are unique to SWI-Prolog, SWISH is
practically not portable to other Prolog
implementations.\footnote{Note that when Pengines is realised through processes
rather than threads, implementing Pengines based on other Prolog systems becomes much
easier.}%

All of these observations lead us to conclude that
an in-browser solution is probably the future of Prolog on the web,
especially for educational use.
The SWISH design is good for offering a collaborative, interactive coding environment and %
(programmatic) %
access to (large) shared data sets,
but its limitations affect the design and capabilities of tools and resources that rely on it. %
When combined with the envisaged authoring suite, both SWISH and in-browser Prolog implementations to come
(such as the Active Logic Documents described elsewhere in this volume \cite{morales2023teaching})
will offer immersive and engaging Prolog teaching as well as learning materials.%

\section*{Acknowledgements}
\ack

\bibliographystyle{splncs04}
\bibliography{paper}

\end{document}